\def\year{2019}\relax
\documentclass[letterpaper]{article} 
\usepackage{aaai19}  
\usepackage{times}  
\usepackage{helvet}  
\usepackage{courier}  
\usepackage{url}  
\usepackage{graphicx}  
\usepackage{amsmath}
\usepackage{diagbox}
\usepackage{color}
\usepackage{amsmath,amssymb,amsfonts,dsfont}
\usepackage{multirow}
\usepackage{subcaption}

\newcommand{\CALD}{\mathcal{D}}

\newcommand{\CALX}{\mathcal{X}}
\newcommand{\CALY}{\mathcal{Y}}

\newcommand{\BBE}{\mathbb{E}}

\newcommand{\expt}[1]{\BBE\left[#1\right]}
\newcommand{\seqt}[1]{\{#1,\, t\geq 0\}}

\newcommand{\commentout}[1]{}
\definecolor{jdk}{rgb}{0.513, 0.012, 0.000}
\definecolor{hjj}{cmyk}{0, 0.5908, 0.4829, 0.1429}

\begin{document}
%
\title{Comparing Sample-wise Learnability Across Deep Neural Network Models}
\author{
Seung-Geon Lee$^1$\thanks{
    This work was performed when Seung-Geon Lee, an undergraduate student, worked as a summer intern at Samsung Research (2018).
}
Jaedeok Kim$^2$\thanks{
	Corresponding author (05jaedeok@gmail.com)
}
Hyun-Joo Jung$^2$
Yoonsuck Choe$^{2,3}$\\
$^1$ Department of Computer Science and Engineering, Seoul National University\\
1 Gwanak-ro, Gwanak-gu,
Seoul, Korea, 08826 \\
$^2$Machine Learning Lab, Artificial Intelligence Center, Samsung Research, Samsung Electronics Co.\\
56 Seongchon-gil, Secho-gu,
Seoul, Korea, 06765 \\
$^{3}$Department of Computer Science and Engineering, Texas A\&M University\\
College Station, TX, 77843, USA
}

\maketitle

\section{Abstract}
\noindent Estimating the relative importance of each sample in a training set has important practical and theoretical value, such as in importance sampling or curriculum learning.
This kind of focus on individual samples invokes the concept of sample-wise learnability: How easy is it to correctly learn each sample (cf.\ PAC learnability)?
\commentout{
(Contrast this with that in the PAC-learning framework where learnability is defined over whole concept classes, not over individual samples.)
}
In this paper, we approach the sample-wise learnability problem within a deep learning context.
We propose a measure of \commentout{sample importance, based on the learnability of the sample} the learnability of a sample with a given deep neural network (DNN) model.
The basic idea is to train the given model on the training set, and for each sample, aggregate the hits and misses over the entire training epochs.
Our experiments show that the sample-wise learnability measure collected this way is highly linearly correlated across different DNN models (ResNet-20, VGG-16, and MobileNet), suggesting that such a measure can provide deep general insights on the data's properties.
We expect our method to help develop better curricula for training, and help us better understand the data itself.

\section{Introduction}
The performance of DNN models depends heavily on the quantity and quality of data.
Furthermore, the order in which the data points are sampled during training makes a big difference in the learning outcome, as shown in latest studies in curriculum learning and self-paced learning \cite{bengio2009curriculum,jiang2015self}.
In this paper, we propose the concept of sample-wise learnability: How easy is it to learn each individual sample, in general, when multiple learning models are considered. Learnability is a well known concept in computational learning theory. However, in the PAC-learning framework for example, learnability is usually defined over a whole concept class, not over individual samples.
We show that sample-wise learnability for a fixed data set, measured using different DNN models, are strongly linearly correlated.
\commentout{
(The correlation is even stronger when the rankings of the learnability are compared.)
}
This way, our approach helps us gain deeper insights into the data itself, and we expect our measure to help automatically generate better curricula for improved performance in DNN training.

\commentout{
Successful training of deep neural networks \textcolor{jdk}{(DNNs)} heavily depends on the quantity and quality of data. Also, the specific order in which the samples are used in training greatly affects the learning outcome.
\commentout{Efficient learning methodologies for deep neural network(DNN) are as important as collecting training samples.}
Curriculum learning and self-Paced learning, inspired by social education in humans, has gain \cite{bengio2009curriculum}.
To determine training order of data samples in curriculum learning, learnability:
Curriculum is determined using a basic order as Easy-to-hard in the human education history.
Estimating the learnability of data samples can give us useful information to develop better curriculum for training, and also selection for coreset training.
Furthermore, It can help us better understand the data.
This paper defines sample-wise learnability of data samples and it is linearly correlated across shows these measures has generality across DNN models by \textcolor{red}{analyzing linear correlation coefficient.}\textcolor{jdk}{[experiments.]}

\textcolor{green}{"The CL methodology has been applied to various applications, the key in which is to find a rank function that assigns learning priorities to training samples" in this study : Self-Paced Curriculum Learning \cite{jiang2015self}
And This paper says \textcolor{red}{"self-paced curriculum"\cite{kumar2010self}} and Curriculum learning both generally start with learning easier aspects of a task,}
}

\begin{figure*}[t]
    \begin{subfigure}[b]{0.30\textwidth}
    \centering
    \includegraphics[width=0.85\linewidth]{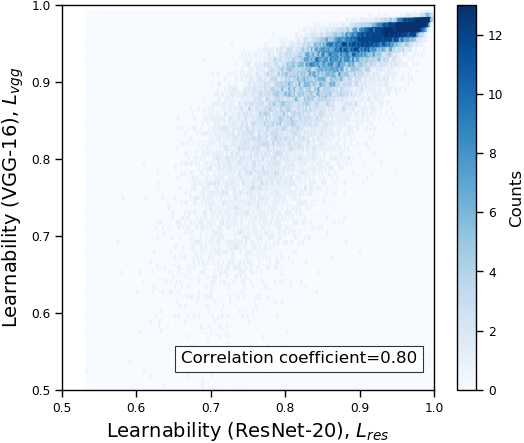}
    \caption{Learnability}
    \label{fig:ra-va}
    \end{subfigure}%
    \begin{subfigure}[b]{0.32\textwidth}
    \centering
    \includegraphics[width=0.85\linewidth]{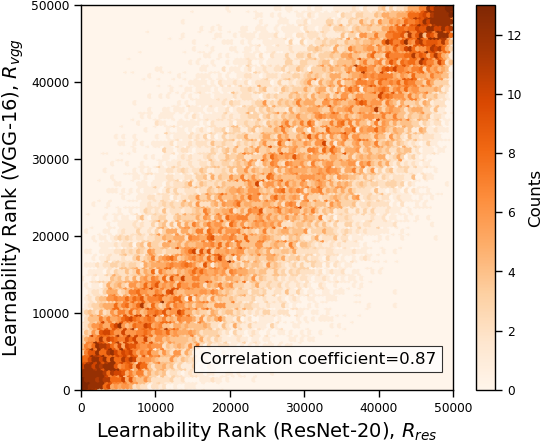}
    \caption{Learnability Rank}
    \label{fig:ra-va_rank}
\end{subfigure}%
\begin{subfigure}[b]{0.35\textwidth}
    \includegraphics[width=0.98\linewidth]{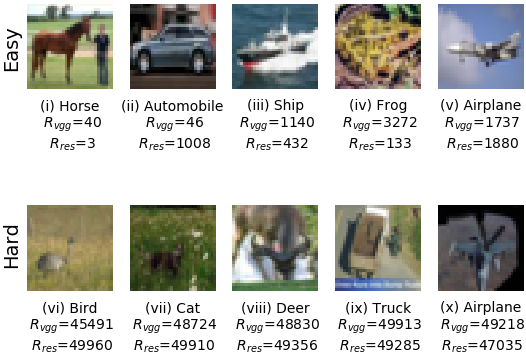}
    \caption{Example Images}
    \label{fig:ind-images}
\end{subfigure}
\caption{
(a) and (b): Distribution (histogram) of samples from the CIFAR-10 data set. The $x$- and $y$-axes correspond to ResNet-20 and VGG-16-based learnability/learnability rank, respectively.
(c): Example images from the CIFAR-10 dataset with their ground truth label and learnability ranks (top row: easy, bottom row: hard).
$R_{vgg}$ and $R_{res}$ represent the learnability rank induced by VGG-16 and ResNet-20 respectively.
Learnability rank 1 means the easiest and 50,000 means the hardest to learn.}
\end{figure*}

\section{Sample-wise Learnability}
\commentout{
Our goal in this paper is to propose a measure that describes how easy to learn from the viewpoint of a single sample.
To this end, we first define the learnability of a sample based on the prediction results by a model over training period.
}

Let $\CALX$ be a domain of inputs and $\CALY:=\{1,\cdots,L\}$ be the set of all possible labels.
A DNN model is a prediction function $f\colon\CALX\to[0, 1]^L$ over $\CALX$, $f(x):=(f_1(x),\cdots,f_L(x))$, such that $\sum_{l=1}^L f_l(x) = 1$ for $x\in\CALX$.
During training, the weights of the DNN model $f$ is updated by an optimizer.
So we denote by $f^{(t)}$ the DNN model after training step $t$.

We take a sample $(X_c, Y_c)$, a pair of input and label, from  $\CALX\times\CALY$ as our reference.
Then $f^{(t)}(X_c)$ is the prediction of $X_c$ by the DNN model after $t$ training steps and $\seqt{f^{(t)}(X_c)}$ can be considered a stochastic process of predictions (by the DNN Model) of the tagged sample $(X_c, Y_c)$ during training.
If the tagged sample $X_c$ is easily learnable, in most training steps a model $f^{(t)}$ should correctly predict the true label $Y_c$ of the tagged input $X_c$.

Based on such an intuition, we define the learnability of an individual sample $X_c$ with respect to a model $f$ as
\begin{align}
    \label{eqn:dft-lbt}
    L_{f}(X_c, Y_c) := \expt{\frac{1}{T} \sum_{t=1}^T f^{(t)}_{Y_c}(X_c)}
\end{align}
where $T$ denotes the total number of training steps.
Although $f^{(t)}_{Y_c}(X_c)$ is the probability that the model predicts the label of $X_c$ as $Y_c$, it is still a random variable since the model $f^{(t)}$ is evolved randomly due to the randomness in the initialization and optimization.
Eq.\ \ref{eqn:dft-lbt} is the expected value over such a quantity $\seqt{f^{(t)}}$.

\commentout{
The value $f^{(t)}_{Y_c}(X_c)$ is the probability of the true label $Y_c$ by the model $f^{(t)}$ after $t$ training steps.
}
So, if $X_c$ is easily learnable, the value of $f^{(t)}_{Y_c}(X_c)$ increases rapidly to 1 as the training step $t$ increases.
Accordingly, the value of $L_{f}(X_c, Y_c)$ also increases.
Otherwise, the value of the probability $f^{(t)}_{Y_c}(X_c)$ remains small and so does the value of $L_{f}(X_c, Y_c)$.
We therefore can say that Eq. \ref{eqn:dft-lbt} faithfully represents the learnability of the sample $(X_c, Y_c)$.

Training of a DNN model is considerably affected by the order in which the samples are drawn and presented to the model, e.g.\ as shown in curriculum learning \cite{jiang2015self}.
So it is also worthy to consider the relative order among training samples in terms of the learnability.

Denoted by $\CALD:=\{(X_1,Y_1),\cdots,(X_N,Y_N)\}$ a training dataset of size $N$ over $\CALX\times\CALY$.
Let $R_{f,i}$ be the learnability rank of the $i$th training sample $(X_i, Y_i)$ in $\CALD$ with respect to the model $f$. Formally, we can write
\begin{align}
	\label{eqn:dft-rank}
    R_{f,i} = \sum_{j=1}^N \textbf{1}_{[L_f(X_i, Y_i) \leq L_f(X_j, Y_j)]}.
\end{align}
Then $L_f(X_i,Y_i)>L_f(X_j,Y_j)$ if $R_{f,i}<R_{f,j}$, which implies learning the $i$th sample is easier than learning the $j$th sample in terms of the learnability.

\section{Experimental Results}

We applied the proposed learnability measure to the CIFAR-10 \commentout{\cite{krizhevsky2014cifar}} data set, using ResNet-20 \cite{he2016deep}, VGG-16 \cite{simonyan2014very}, and MobileNet \cite{howard2017mobilenets}.
To compare the learnability of each sample with respect to different models, we used the same training options for all models.
In our experiment we considered a single training epoch as a training step and used $T=200$.


We plot the learnability of samples with respect to the VGG-16 and ResNet-20 in Figure \ref{fig:ra-va}.
As we can see in the figure, the learnability of both models are positively correlated, and
the correlation coefficient is 0.80.
Figure \ref{fig:ra-va_rank} shows the relation of learnability rank induced by VGG-16 and that induced by ResNet-20.
Similar with the case of learnability, the learnability rank of samples are also positively correlated (correlation coefficient = 0.87).

Figure \ref{fig:ind-images} shows actual examples from the CIFAR-10 training set (the set includes a total of 50,000 images).
The images in the top row have high rank (small learnability rank value) which means that they are easy to learn.
As we can see in the figure, the images in the top row have well defined features and we can easily classify them.
In contrast, the images in the bottom row have low rank (large learnability rank value) and hard to classify even for humans. For example, scale is too small (Figure \ref{fig:ind-images} (vi) and (vii)) or viewpoint is atypical (Figure \ref{fig:ind-images} (ix) and (x)).

\begin{table}[t]
  \caption{
  Correlation across models.
  The correlation coefficients of learnability and that of learnability rank (parenthesized) are shown. Note: correlation matrices are symmetric, so redundant
  information was omitted.
  \commentout{
  	Because the correlation coefficient is symmetric, we show the correlation coefficient of learnability at the upper triangular matrix and the correlation coefficient of rank is represented at the lower triangular matrix with parenthesis.
    }
  }
  \label{tbl:correlation}
  \centering
  \begin{tabular}{c|lclclcl}
    \hline
       & VGG-16 & ResNet-20 & MobileNet \\
    \hline
    VGG-16       & -           & 0.796 (0.867)   & 0.713 (0.792)   \\
    ResNet-20    & - & -          & 0.774 (0.782)  \\
    MobileNet    & - & - &    -        \\
    \hline
  \end{tabular}
\end{table}

The full comparison across all tested models in summarized in Table\ \ref{tbl:correlation}.
The correlation coefficients in all cases are higher than 0.71.
The results suggest that our proposed learnability and rank are consistent across models.

From the above results, we can argue that the proposed sample-wise learnability is an effective measure to estimate the  importance of individual samples in a given training set.

\section{Conclusion}
In this paper, we introduced the concept of sample-wise learnability (and it's rank-based variant) based on the prediction performance during training.
We experimentally showed that the sample-wise learnability (and its rank) for a given data set is linearly correlated across different models.
We expect our measure to help develop better curricula for training, and help us better understand the data itself.


\bibliographystyle{aaai}
\bibliography{main}

\begin{thebibliography}{}

\bibitem[\protect\citeauthoryear{Bengio \bgroup et al\mbox.\egroup
  }{2009}]{bengio2009curriculum}
Bengio, Y.; Louradour, J.; Collobert, R.; and Weston, J.
\newblock 2009.
\newblock Curriculum learning.
\newblock In {\em ICML},  41--48.
\newblock ACM.

\bibitem[\protect\citeauthoryear{He \bgroup et al\mbox.\egroup
  }{2016}]{he2016deep}
He, K.; Zhang, X.; Ren, S.; and Sun, J.
\newblock 2016.
\newblock Deep residual learning for image recognition.
\newblock In {\em CVPR},  770--778.

\bibitem[\protect\citeauthoryear{Howard \bgroup et al\mbox.\egroup
  }{2017}]{howard2017mobilenets}
Howard, A.~G.; Zhu, M.; Chen, B.; Kalenichenko, D.; Wang, W.; Weyand, T.;
  Andreetto, M.; and Adam, H.
\newblock 2017.
\newblock Mobilenets: Efficient convolutional neural networks for mobile vision
  applications.
\newblock {\em arXiv preprint arXiv:1704.04861}.

\bibitem[\protect\citeauthoryear{Jiang \bgroup et al\mbox.\egroup
  }{2015}]{jiang2015self}
Jiang, L.; Meng, D.; Zhao, Q.; Shan, S.; and Hauptmann, A.~G.
\newblock 2015.
\newblock Self-paced curriculum learning.
\newblock In {\em AAAI}, volume~2, ~6.

\bibitem[\protect\citeauthoryear{Simonyan and
  Zisserman}{2014}]{simonyan2014very}
Simonyan, K., and Zisserman, A.
\newblock 2014.
\newblock Very deep convolutional networks for large-scale image recognition.
\newblock {\em arXiv preprint arXiv:1409.1556}.

\end{thebibliography}


\begin{thebibliography}{}

\bibitem[\protect\citeauthoryear{Chollet and others}{2015}]{chollet2015keras}
Chollet, F., et~al.
\newblock 2015.
\newblock Keras.
\newblock \url{https://keras.io}.

\bibitem[\protect\citeauthoryear{Ruder}{2016}]{ruder2016overview}
Ruder, S.
\newblock 2016.
\newblock An overview of gradient descent optimization algorithms.
\newblock {\em arXiv preprint arXiv:1609.04747}.

\end{thebibliography}

\end{document}


%
\title{Supplementary Material: Sample-wise Learnability Across Deep Neural Network Models}

\author{
Seung-Geon Lee$^1$
Jaedeok Kim$^2$
Hyun-Joo Jung$^2$
Yoonsuck Choe$^{2,3}$\\
$^1$ Department of Computer Science and Engineering, Seoul National University\\
1 Gwanak-ro, Gwanak-gu,
Seoul, Korea, 08826 \\
$^2$Machine Learning Lab, Artificial Intelligence Center, Samsung Research, Samsung Electronics Co.\\
56 Seongchon-gil, Secho-gu,
Seoul, Korea, 06765 \\
$^{3}$Department of Computer Science and Engineering, Texas A\&M University\\
College Station, TX, 77843, USA
}
\maketitle


\section{Correlation With Small Size DNN}

\begin{figure*}[t]
  \begin{subfigure}[b]{0.5\textwidth}
    \centering
    \includegraphics[width=0.8\linewidth]{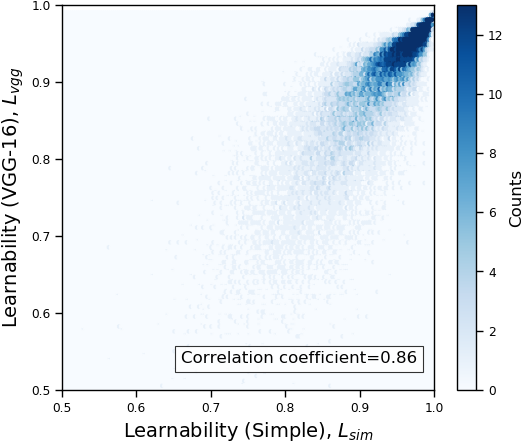}
    \caption{Learnability between Simple and VGG-16}
  	\label{fig:sim-vgg}
  \end{subfigure}%
  \begin{subfigure}[b]{0.53\textwidth}
    \centering
    \includegraphics[width=0.8\linewidth]{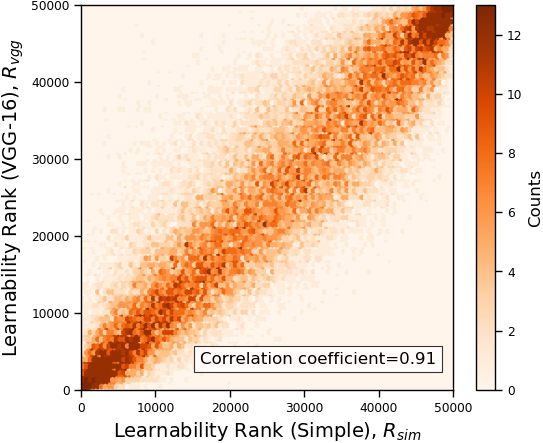}
    \caption{Learnability rank between Simple and VGG-16}
    \label{fig:sim-vgg_rank}
  \end{subfigure}

  \begin{subfigure}[b]{0.5\textwidth}
    \centering
    \includegraphics[width=0.8\linewidth]{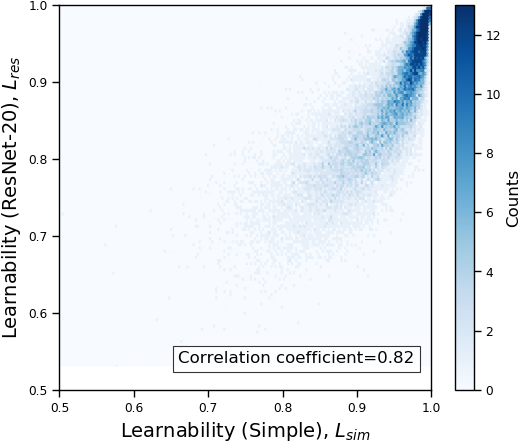}
    \caption{Learnability between Simple and ResNet-20}
  	\label{fig:sim-vgg}
  \end{subfigure}%
  \begin{subfigure}[b]{0.53\textwidth}
    \centering
    \includegraphics[width=0.8\linewidth]{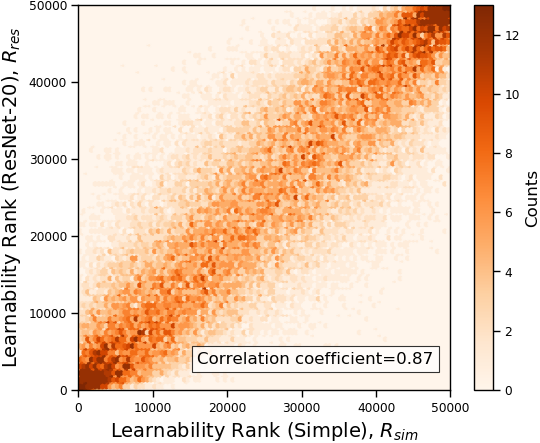}
    \caption{Learnability rank between Simple and ResNet-20}
    \label{fig:sim-vgg_rank}
  \end{subfigure}
\caption{
2D histogram of samples in CIFAR-10 training set.
We divided each axis into 200 bins in (a) and (c) and 100 bins in (b) and (d).
}
\label{fig:corr-2d-hist}
\end{figure*}

The models used in our experiments (ResNet-20, VGG-16, and MobileNet) might be over-capable to CIFAR-10 dataset.
So we also investigated whether there is positive correlation in case of a simple model.
In our experiment we used a small size model consisting of 4 convolution layers and 2 fully connected layers which is implemented in Keras example \cite{chollet2015keras}.
To compare the learnability of each sample with respect to different models, we used the same training options as before, i.e. $T=200$.

Figure \ref{fig:corr-2d-hist} shows the relation of learnability and rank via 2D histogram across models.
As we can see, even if we used a simple model, we can validate our claim that our proposed measure has a positive correlation across models.

\section{Correlation Across Optimizers}

\begin{figure*}[t]
  \begin{subfigure}[b]{0.5\textwidth}
    \centering
    \includegraphics[width=0.8\linewidth]{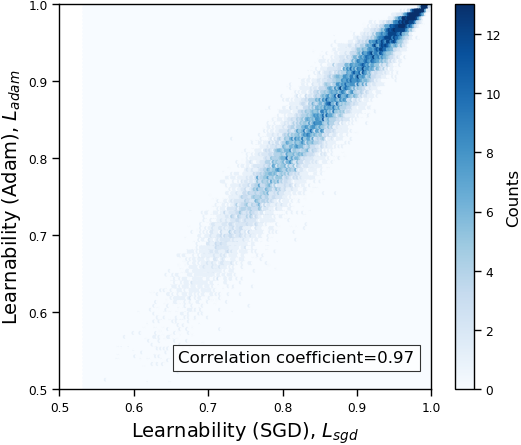}
    \caption{Learnability}
  	\label{fig:sim-vgg}
  \end{subfigure}%
  \begin{subfigure}[b]{0.53\textwidth}
    \centering
    \includegraphics[width=0.8\linewidth]{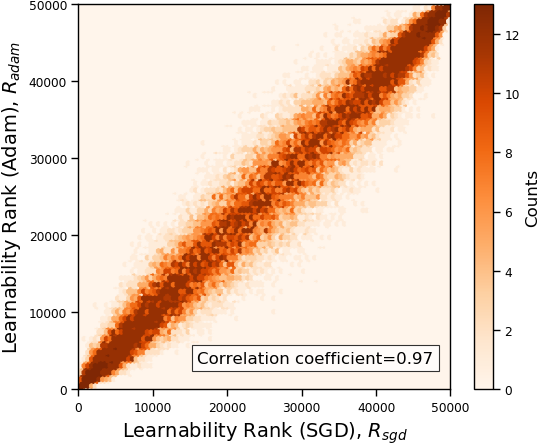}
    \caption{Learnability Rank}
    \label{fig:sim-vgg_rank}
  \end{subfigure}
\caption{
(a) and (b): 2D histogram of samples in CIFAR-10 training set. $x$- and $y$-axis are corresponding to SGD and Adam, respectively. We divided each axis into 200 bins and 100 bins, respectively.
}
\label{fig:corr-opt}
\end{figure*}

\begin{table}[t]
  \caption{
  	Correlation coefficient across models in both learnability and rank.
    The upper triangular matrix shows the correlation coefficient of learnability and the lower triangular matrix shows the correlation coefficient of rank with parenthesis.}
  \label{tbl:corr-across-opt}
  \centering
  \begin{tabular}{c|c c c}
    \hline
             & SGD       & Adam     & RMSprop  \\
    \hline
    SGD      & -         & 0.9698   & 0.9742   \\
    Adam     & (0.9730)  & -        & 0.9932   \\
    RMSprop  & (0.9756)  & (0.9936) &    -     \\
    \hline
  \end{tabular}
\end{table}

The learnability is probably influenced by a training method such as a selection of optimizer.
An optimizer used in training affects how to evolve the model $f^{(t)}$.
So a sequence $\seqt{f^{(t)}}$ of trained DNN models depends on training steps.
Consequently, using a different optimizer could make the learnability of a sample different as well as using different architecture.

In order to investigate the effect of an optimizer on learnability, we also compare different models having the same architecture but trained by different optimization algorithms.
In our experiment, we trained ResNet-20 by using three different optimization algorithms: Stochastic gradient descent (SGD), Adam, and RMSprop optimizer \cite{ruder2016overview}.
At each optimizer the learning rate was set by 0.01, 0.001, 0.001, respectively.

The relation between the learnability induced by SGD and Adam is shown in Figure \ref{fig:corr-opt}.
In addition, Table \ref{tbl:corr-across-opt} shows the correlation coefficient for all possible combinations.
Similar to the previous experiments, we can find a linear correlation in both learnability and rank although we change the optimization algorithms.

\bibliographystyle{aaai}
\bibliography{supp}